\documentclass[letterpaper, 10 pt, conference]{ieeeconf}  

\IEEEoverridecommandlockouts                              

\usepackage{graphics} 
\usepackage{epsfig} 
\usepackage{mathptmx} 
\usepackage{amsmath} 
\usepackage{amssymb}  
\usepackage{cite} 
\usepackage{float}
\usepackage[linesnumbered,ruled,vlined]{algorithm2e}

\usepackage[dvipsnames]{xcolor}

\newcommand{\erhan}[1]{\textcolor{black}{#1}}
\newcommand{\edit}[1]{\textcolor{black}{#1}}

\title{\LARGE \bf
Context-Based Echo State Networks with Prediction Confidence for Human-Robot Shared Control

}

\usepackage{authblk}

\author[1]{%
{\hspace{1mm}Negin Amirshirzad\thanks{\texttt{negin.amirshirzad@ozu.edu.tr}}}
}%
\author[1]{%
{\hspace{1mm}Mehmet Arda Eren\thanks{\texttt{arda.eren@ozu.edu.tr}}}
}%
\author[1,2]{%
	{\hspace{1mm}Erhan Oztop\thanks{\texttt{erhan.oztop@otri.osaka-u.ac.jp}}}%
}
\affil[1]{Ozyegin University, Istanbul, Turkey}
\affil[2]{Osaka University, Osaka, Japan}

\begin{document}

\maketitle
\thispagestyle{empty}
\pagestyle{empty}

\begin{abstract}

In this paper, we propose a novel lightweight learning from demonstration (LfD) model based on reservoir computing that can learn and generate multiple movement trajectories with prediction intervals, which we call as Context-based Echo State Network with prediction confidence (CESN+). CESN+ can generate movement trajectories that may go beyond the initial LfD training based on a desired set of conditions while providing confidence on its generated output. To assess the abilities of CESN+, we first evaluate its performance against Conditional Neural Movement Primitives (CNMP), a comparable framework that uses a conditional neural process to generate movement primitives. Our findings indicate that CESN+ not only outperforms CNMP but is also faster to train and demonstrates impressive performance in generating trajectories for extrapolation cases. 
In  human-robot shared control applications, the confidence of the machine generated trajectory is a key indicator of how to arbitrate control sharing. To show the usability of the CESN+ for human-robot adaptive shared control, we have designed a proof-of-concept human-robot shared control task and tested its efficacy in  adapting the sharing weight between the human and the robot by comparing it to a fixed-weight control scheme. The simulation experiments show that with CESN+ based adaptive sharing the  total human load in shared control can be significantly reduced. Overall, the developed CESN+ model is a strong lightweight LfD system with desirable properties such fast training and ability to extrapolate to the new task parameters while producing robust prediction intervals for its output.

\end{abstract}

\section{INTRODUCTION}

Estimating the confidence of predictions in machine learning and robotics is paramount to ensuring the reliability and safety of autonomous systems. In complex real-world scenarios, uncertainties abound, ranging from noisy sensor data to inherent ambiguity in human interactions.  By quantifying the confidence of the predictions, these systems can make more informed decisions, especially in high-risk environments. For instance, in self-driving cars, accurately estimating the confidence of object detection can prevent accidents by indicating when the system is uncertain about its surroundings, prompting a cautious response or seeking human intervention. Moreover, in robotics applications like medical diagnosis or industrial automation, understanding prediction confidence allows for robust decision-making, reducing the risk of costly errors or adverse outcomes.

\edit{The integration of confidence in prediction within autonomous systems is particularly critical in human-robot shared control settings, where both human and robot contribute to decision-making and action execution. In these systems, accurately estimating the reliability of predictions is essential to ensure smooth, safe collaboration and to mitigate potential risks. Furthermore, adaptive control based on confidence levels can reduce the likelihood of collisions or operational conflicts, thereby enhancing overall system safety and performance \cite{nikolaidis2017human}.
Implementing confidence-aware models also addresses the problem of "overconfident" predictions, which could otherwise lead to incorrect robot behaviors or misinterpretations by human operators \cite{9926657} Techniques that monitor and adjust robot actions based on real-time confidence in human predictions have been shown to improve operational safety in shared environments, balancing the trade-off between responsiveness and caution \cite{9926657, 9812048,fisac2018probabilistically}. Such frameworks that utilize probabilistic safety methods and models of human behavior are increasingly being explored for adaptive, shared control in applications such as assistive robotics and autonomous driving.}

At its core, reservoir computing consists of a fixed, randomly generated recurrent neural network, known as the reservoir, coupled with a trainable readout layer \cite{schrauwen2007overview}. Unlike traditional neural networks where all parameters are trained, in reservoir computing, only the readout layer is optimized while the reservoir's internal connections remain fixed. This design simplifies training and accelerates learning.
Reservoir computing serves as the foundational framework for Echo State Networks (ESNs) \cite{jaeger2007echo}. ESNs are characterized by a reservoir with a spectral radius typically less than or close to one, ensuring stability and dynamic richness. They also employ discrete-time update rules, simplifying their implementation and enabling efficient processing of temporal data. These reservoir neurons serve as a dynamic memory system that processes information from input data. 

In this work, we extend our Learning from Demonstration (LfD) model, the Context-based Echo State Network (CESN), to incorporate prediction confidence, introducing the Context-based Echo State Network with prediction confidence (CESN+). This enhanced model generates movement trajectories conditioned on desired human movements while providing a measure of confidence in its predictions.

We first evaluate CESN+ against a comparable framework, Conditional Neural Movement Primitives (CNMP) to assess its performance in generating robotic movement primitives given a condition, and then apply CESN+ in a human-robot shared control experiment and use its confidence of prediction to obtain an adaptive weight parameter for a dynamic share of control between the human and robot.\\
The rest of the paper is structured as follows: Section II discusses related work in the field, Section III details our methodology, section IV compares our approach with CNMP, Section V presents our robotic experiments, and Section VI concludes our paper.

\section{Related Work}

\subsection{Learning from Demonstrations}
Learning from demonstration has been applied to various robotic tasks where the robot can either mimic the motion of the demonstrator or learn how the demonstrator acts, reacts, and handles errors \cite{argall2009survey}. LfD can implicitly capture the task requirements and constraints from the demonstrator, enabling adaptive behavior. 
Conditional Neural Processes (CNPs) \cite{garnelo2018conditional} have the benefit of fitting observations efficiently with linear complexity in the number of context input-output pairs and can learn predictive distributions conditioned on context sets. However, a fundamental drawback of CNPs is that they tend to underfit \cite{bruinsma2023autoregressive}, less expressive predictive distributions.

Conditional Neural Movement Primitives (CNMPs) \cite{seker2019conditional} are a framework built on top of CNPs \cite{garnelo2018conditional}, designed for robotic movement learning and generation. CNMPs can be conditioned on single or multiple time-steps to produce trajectory distributions that satisfy any number of given conditions.
Context-based Echo State Networks (CESNs) were introduced in our previous work \cite{amirshirzad2023context} as a lightweight framework for generating robotic movement primitives. The linear read-out weights in CESNs capture context-dependent dynamics, allowing for the generation of different movement patterns that satisfy the given condition provided by the context.

\subsection{Adaptive Shared Control}
Adaptive shared control involves dynamically adjusting the control weight between a human operator and a robotic system based on real-time data. As the most well-known example of adaptive shared control, autonomous vehicles utilize shared control systems to enhance safety and comfort by adapting the level of autonomy according to driver state and environmental conditions, leading to significant performance improvements \cite{huang2021human}.
Human actions are typically goal-directed \cite{csibra2003teleological}, making accurate intention prediction challenging due to the diverse strategies employed by different operators and the influence of experience on movement patterns \cite{amirshirzad2019human}. A comprehensive review of intention estimation techniques categorizes these methods into explicit and implicit approaches \cite{argall2020}, with implicit methods inferring intentions from observable behaviors rather than direct communication. Recent advancements have incorporated machine learning to enhance real-time predictions of user intent based on motion sensory data \cite{li2013human}.
While visual cues are commonly used for intention estimation, they can be insufficient, highlighting the need for a combination of multiple data sources to achieve greater reliability \cite{butepage2017deep}. Wearable sensing devices also offer opportunities to capture physiological data for intention inference, though scalability poses challenges in applying these methods to larger systems \cite{wang2018human,losey2018review}. By leveraging intention estimation techniques, shared control systems can better integrate the strengths of both humans and robots, ultimately enhancing overall performance while decreasing the cognitive burden on human operators \cite{franchi2012shared,boessenkool2012task}.

\subsection{Uncertainty in Machine Learning Models}

Uncertainty in machine learning reflects the degree of confidence a model has in its predictions, helping to identify potential errors or variability caused by limitations in the model or the data \cite{gawlikowski2023survey}. The uncertainty caused by the data, known as aleatoric uncertainty \cite{hullermeier2021aleatoric, wang2019aleatoric, bae2021estimating}, is often due to the presence of noise or randomness in data generation, which cannot be reduced through training. On the other hand, model (epistemic) uncertainty \cite{hullermeier2021aleatoric, lahlou2021deup, nguyen2019epistemic} arises from limited training data, or low model complexity, and can, in principle, be reduced.

Quantifying uncertainty can be beneficial in robotics, especially in making reliable decisions in dynamic and unpredictable environments, allowing robots to assess risks and adapt to uncertain situations more effectively \cite{kok2020trust, lauri2022partially, lanillos2021active}.

\edit{Various methods have been developed to measure and quantify uncertainty in machine learning. Gaussian Processes (GPs) \cite{rasmussen2003gaussian} are non-parametric models that quantify uncertainty by providing a probability distribution for each input point in the feature space.}
Although GPs are reliable models in terms of uncertainty quantification, they scale poorly with large datasets or high-dimensional input spaces \cite{liu2020gaussian}.
Consequently, to address these scalability issues, several neural network-based methods have been developed \cite{oberdiek2018classification, cavalcanti2016combining, amini2020deep, bayer2013training, goan2020bayesian, gal2015dropout} including the CNP \cite{garnelo2018conditional} and CNMP \cite{seker2019conditional} models reviewed above.
For example, Bayesian neural networks \cite{goan2020bayesian} and Monte Carlo dropout \cite{gal2015dropout} are commonly used models that capture epistemic uncertainty by defining probabilistic distributions over model parameters. However, these methods can be computationally expensive and require specialized models or inference techniques to be deployed directly for robotic applications.

Unlike the aforementioned methods, the prediction intervals \cite{heskes1996practical} offer a more direct approach by providing the uncertainty in estimating the mean response and the variability of individual observations around that mean, under the condition of a linear dependency between the input and output variables. As opposed to models relying on Bayesian inference, prediction intervals can be retrieved using frequentist methods such as quantile regression \cite{taylor1999quantile, zarnani2019quantile}, bootstrapping \cite{pan2016bootstrap}, or conformal prediction \cite{xu2021conformal}. 
This characteristic of the method offers a simpler interpretation of uncertainty, as no probabilistic model is assumed over the samples, which makes it more accessible to robotic applications.

\section{Methodology}

\subsection{Conditional Neural Movement Primitives}
Conditional neural movement primitives (CNMP) \cite{seker2019conditional} is a framework designed to learn and generate robotic movement primitives. CNMPs can be conditioned on single or multiple time-steps to produce trajectory distributions to satisfy the given conditions.
Having a set of inputs (observations) $O{(x_i,y_i)}$, and a set of targets $T{x_i}$ we would like to learn a conditional distribution $Q( f(T) | O, T )$ over all the functions $f :X \rightarrow Y$.
Encoder, which is a multi-layer perceptron (MLP) with ReLU activation function, takes some $(x, y)$ pairs as context points, and results in an individual representation for each pair as in equation (\ref{encoder}). 
\begin{equation}
    r_i = E (x_i,y_i) \quad 
    \forall (x_i,y_i) \in O
    \label{encoder}
\end{equation}
Aggregator combines all the individual representations and results in a general representation which contains the information about the underlying unknown function that maps the inputs to outputs as in equation (\ref {agre}).
\begin{equation}
    r= r_1 	\oplus r_2 	\oplus ... 	\oplus r_n            
    \label{agre}
\end{equation}
Decoder which is also an MLP with ReLU activation function takes the general representation and some $x_t$ values as targets, and results in the mean and variance of the estimated output as in equation (\ref{dec}).
\begin{equation}
    (\mu _t , \sigma _t^2) = D(x_t , r) \quad
    \forall (x_t) \in T
    \label{dec}
\end{equation}
One challenge faced CNMPs is their difficulty with extrapolation, where the accuracy of predictions significantly decreases when the model encounters unseen inputs \cite{akbulut2020acnmp}.

\subsection{Context based echo state networks}


\edit{The task of a typical Echo State Network (ESN) is to learn a target output sequence given a time-varying input sequence. ESNs are a type of recurrent neural network (RNN) that leverages a unique reservoir of randomly connected neurons to process temporal data.} The weights of the input layer and the reservoir are fixed.
These weights are initialized randomly \cite{lukovsevivcius2012practical}.
$u(t) \in \mathbb{R}^{N_u}$ denotes the input signal, $x(t) \in \mathbb{R}^{N_x}$ denotes the state of the reservoir, and $y(t) \in \mathbb{R}^{N_y} $ denotes the output value. Here $ t = 1,..., N $ is the discrete-time and $N$ is the number of data points in the training dataset. 
The task is to learn a model with output $y(t)$  , where $y(t)$ matches $y_{target}(n)$ minimizing
an error measure $E(y, y_{target})$. The typical
update equations for the state of the reservoir with leaky integration are given as follows: 
\begin{equation}
    \label{eq1}
    \tilde {x}(t) = tanh (W^{in}[1; u(t)] + W x(t-1)) 
\end{equation}

\begin{equation}
    \label{eq2}
    x(t) = (1 - \alpha)x(t-1) + \alpha \tilde{x}(t),
\end{equation}
where $\alpha$ is the leaking rate, and $ \tilde {x} (t) \in \mathbb{R}^{N_x} $ is the update, and $W^{in} \in \mathbb{R}^{N_x \times (N_u+1)}$ and $W \in \mathbb{R}^{N_x \times N_x}$ are the input and reservoir weight matrices respectively.
The output value is typically obtained by \erhan{equation} (\ref{eq3}), however, in our implementation, we defined the readout \erhan{equation as in} (\ref{eq4}) to ensure that regression is based purely on the reservoir dynamics.
\begin{equation}
    \label{eq3}
    y(t) = W^{out}[1; u(t); x(t)]
\end{equation}

\begin{equation}
    \label{eq4}
    y(t) = W^{out}[1; x(t)]
\end{equation}
Where $W^{out} \in \mathbb{R}^{N_y \times (N_x + 1) } $ (based on (\ref{eq4})) is the output weight matrix and can be computed by linear regression.

With the objective of extending the ESN application from single-sequence learning to multiple-sequence learning, in our previous work, we proposed inserting a context in the input signal to alter the reservoir dynamics  and estimate the output accordingly. We call these Context-based Echo State Networks (CESNs) \cite{amirshirzad2023context}. In CESN training, the network receives multiple movement trajectories, each associated with a specific context. This context can define the movement's objective, such as specifying a target position that the movement should reach or identifying an obstacle position that the movement should avoid. This approach allows CESNs to flexibly learn and generate movement primitives based on the provided context. While CESNs excel in generalization, computational efficiency, and generating context-dependent movements, they lack the ability to quantify the quality or reliability of their predictions.

\begin{figure*}[ht] 
\centering
\includegraphics[width=0.9\textwidth]{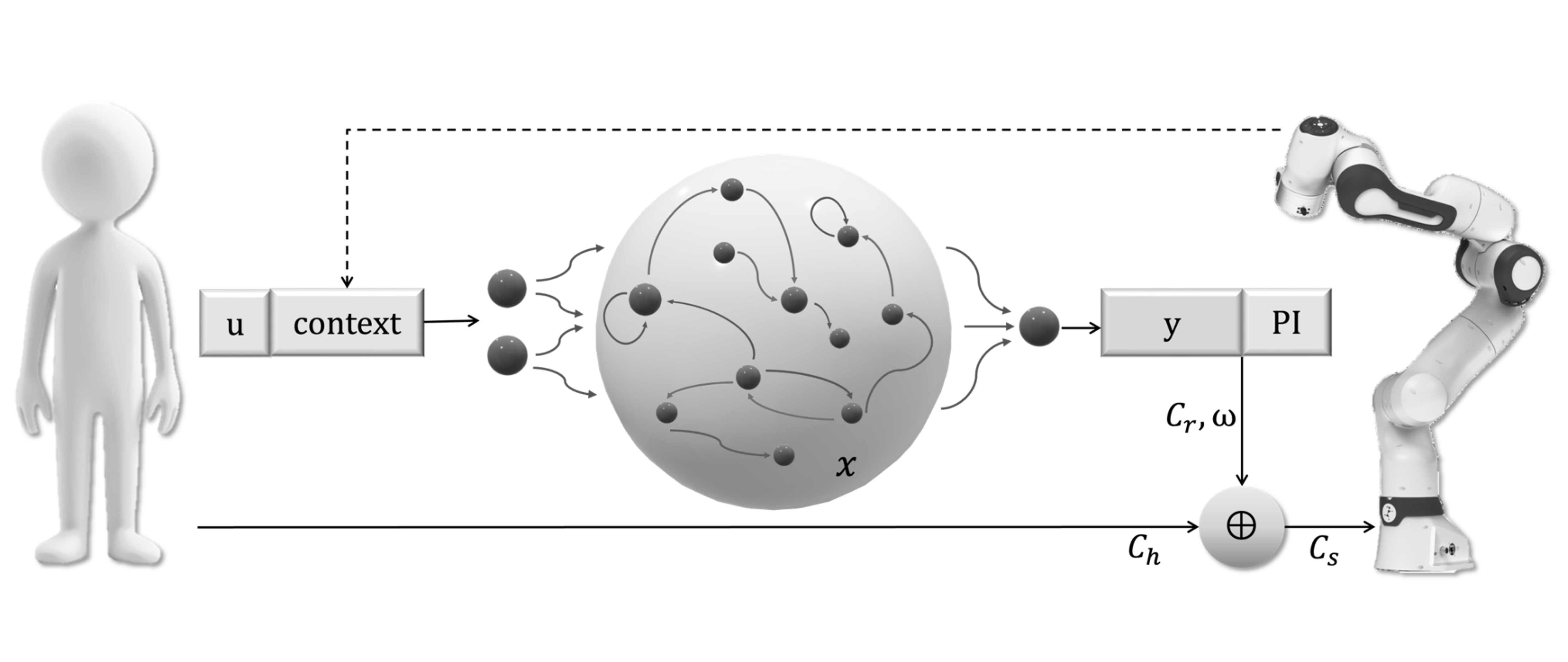}
\vspace{-5mm}
\caption{Human-Robot shared control with Context based echo state network. The shared control command is obtained via the weighted convex combination of human and robot commands. The robot command and weighting factor are obtained from the CESN+. \edit{During task execution, the robot state can be fed-back to the reservoir continuously or the state can be used as the context at discrete time points to condition the prediction at desired time points.}}
\label{fig:esn}

\end{figure*}

To address this limitation, we introduce CESN+, which integrates a prediction confidence measure by utilizing the Prediction Interval (PI) framework from statistical regression. This enhancement enables CESN+ not only to generate movement trajectories but also to provide a quantifiable confidence level for its predictions. The CESN+ architecture is shown in Fig. \ref{fig:esn}.

\subsection{Prediction Interval}
In linear regression, a prediction interval provides a range of values within which future observations are expected to fall with a certain level of confidence. While the more commonly known confidence intervals focus on estimating the mean value of the response variable at a given predictor value, prediction intervals provide a broader range accounting for both the uncertainty in estimating the mean response and the variability of individual observations around that mean \edit{\cite{heskes1996practical}.}


The prediction interval takes into account the uncertainty associated with both the estimated regression coefficients and the variability of the response variable around the regression line. This involves incorporating the standard error of the regression estimate, along with the residual standard error, which captures the scatter of data points around the regression line, and is calculated as in equation (\ref{eq:pi}) \edit{\cite{dybowski2001confidence}:}


\begin{equation}
\label{eq:pi}
\hat{Y}_{pred} \pm t_{\alpha/2} \times s \times \sqrt{1 + \frac{1}{n} + \frac{(X_{pred} - \bar{X})^2}{\sum_{i=1}^{n}(X_i - \bar{X})^2}}
\end{equation}


\edit{Here, \(\hat{Y}_{pred}\) represents the predicted value for the new observation, \(t_{\alpha/2}\) is the critical value of the t distribution, and a significance level of \(\alpha/2\), where \(\alpha\) is the desired confidence level, \(s\) denotes the residual standard error of training phase, capturing the scatter of data points around the regression line, \(X_{pred}\) is the predictor value which is the reservoir state for the new observation, \(\bar{X}\) is the mean of all predictor values, \(X_i\) are the observed predictor values in the training dataset, \(n\) is the number of observations in the dataset.}

When X becomes multidimensional, the formula for the prediction interval in linear regression adjusts accordingly as in equation (\ref{eq:pim}). 

\begin{equation}
\label{eq:pim}
 \hat{Y}_{pred} \pm t_{\alpha/2} \times s \times \sqrt{1 + \mathbf{X}_{pred}^T (\mathbf{X}^T \mathbf{X})^{-1} \mathbf{X}_{pred}}   
\end{equation}

\edit{In this formula, \(\mathbf{X}\) is a matrix made of concatenation of \(\mathbf{x}\)'s, where each \(\mathbf{x}\) is the reservoir state for a demonstration in the training dataset. \(\mathbf{X}_{pred}\) represents the predictor vector that is the reservoir state for the new observation. \(t_{\alpha/2}\), \(s\), and \(\hat{Y}_{pred}\) retain the same meanings as in the previous equation (\ref{eq:pi}).} 



\section{Emulating CNMP with CESN+}
To evaluate and compare the performance of Context-based Echo State Networks with prediction confidence (CESN+) and Conditional Neural Movement Primitives (CNMPs), four simple trajectories were used as demonstration data for both models. In the CNMP architecture, the observation layers have sizes of [128, 128, 128], and the decoder layers are sized [128, 128, 2]. These parameter choices follow those used in prior work on CNMPs \cite{amirshirzad2022adaptive}, as they balance the expressiveness of the network with computational efficiency. For CESN+, a reservoir size of 500 was used. Although CNMPs require a significant amount of time for training, CESN+ offers the advantage of nearly instantaneous training. In this evaluation, the models were provided with a single conditioning point (the red dot) as input. The task was to generate trajectories that meet the given condition(s), while providing a measure of their confidence in their prediction.

CNMPs can be conditioned with any number of context points as previously mentioned. In contrast, for CESN+, the specific number of context points needs to be predefined prior to training. The confidence in CNMP estimations is represented by the standard deviation of the predicted normal distribution, on the other hand, CESN+ confidence is indicated through a prediction interval.
It is important to recognize that the absolute values of these confidence measures may not always be directly comparable. Nevertheless, they can give a relative indication of the confidence in the estimation within their respective input ranges.

\subsection{Single context point}

In the first comparison, we evaluated the generated trajectories of the two models when given a single conditioning point. Both models were trained on four simple movement trajectories. This comparison aimed to focus on the accuracy of the generated trajectory in relation to the given condition.

\begin{figure}[ht!]
    \centering
    \includegraphics[width=1\linewidth]{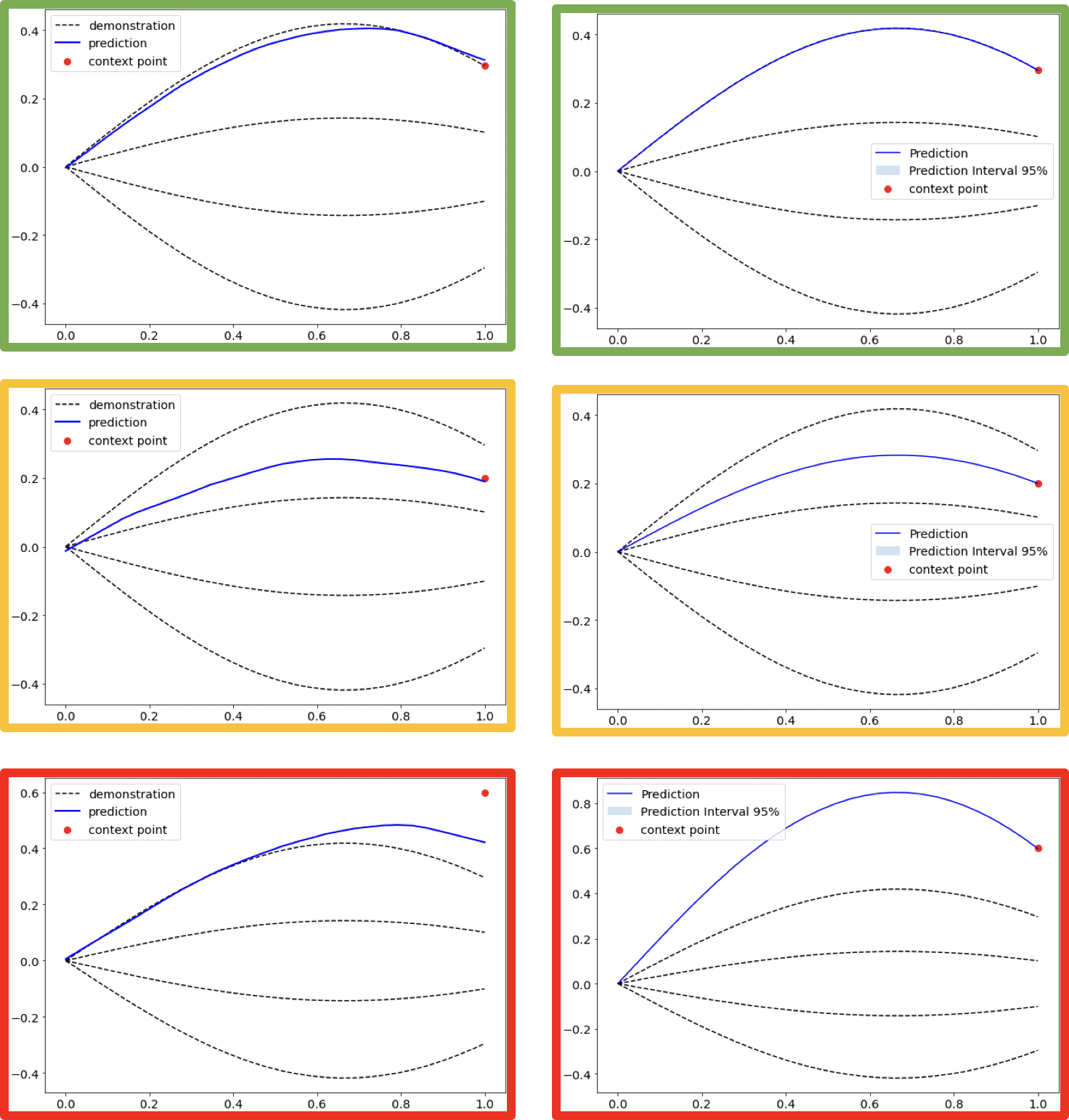}
    \caption{CNMP vs. CESN+ Predictions: The left plots display CNMP predictions, while the right plots illustrate CESN+ predictions for scenarios involving known (green), interpolation (yellow), and extrapolation (red) cases, each with one context point.}
    \label{fig:cvc}
    \vspace{-4mm}
\end{figure}

The resulting trajectories, illustrated in Figure \ref{fig:cvc}, showcase three types of predictions: green trail, representing known paths observed during training; yellow depicting interpolated paths between the training data; and red trajectories, representing extrapolated paths beyond the demonstrated cases. These results clearly indicate that CESN+ exhibits superior performance in generating accurate and smooth extrapolated trajectories, effectively handling scenarios that go beyond the training data. This demonstrates its robustness and generalization capabilities compared to CNMPs.

We performed a similar task to further evaluate the models' performance under more complex scenarios and to assess the accuracy of the generated trajectory relative to the provided context, as well as the reliability of each model's confidence metric.

\begin{figure}[htbp!]
    \centering
    
    \includegraphics[width=\linewidth]{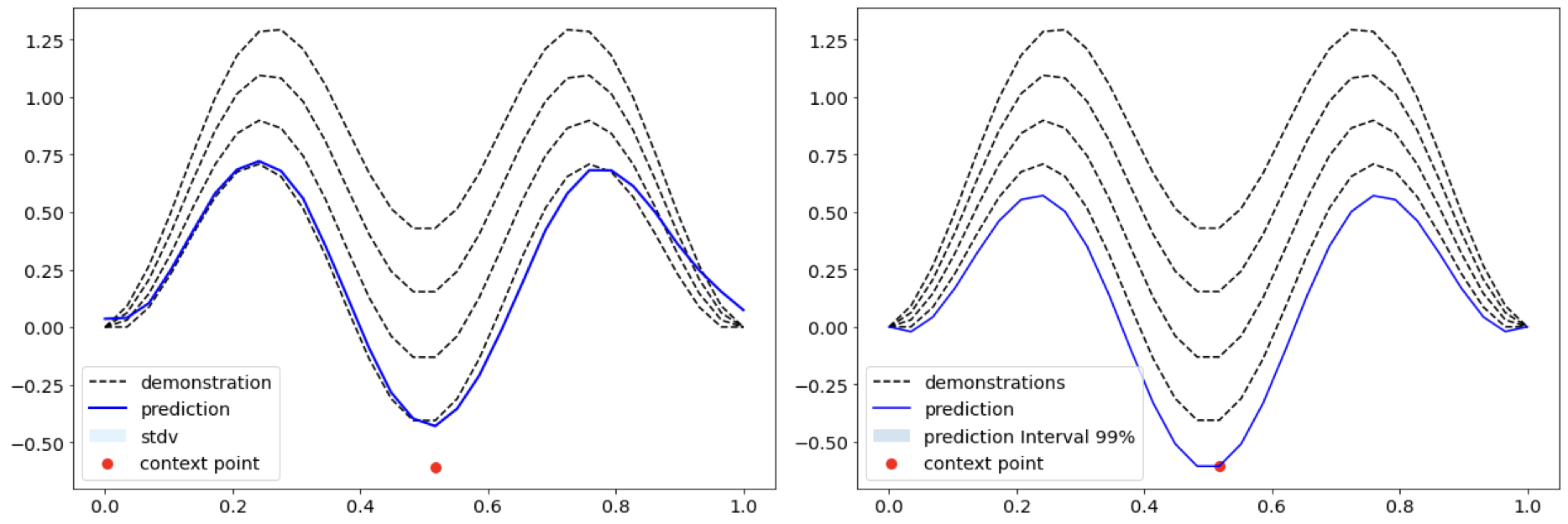}
    \includegraphics[width=\linewidth]{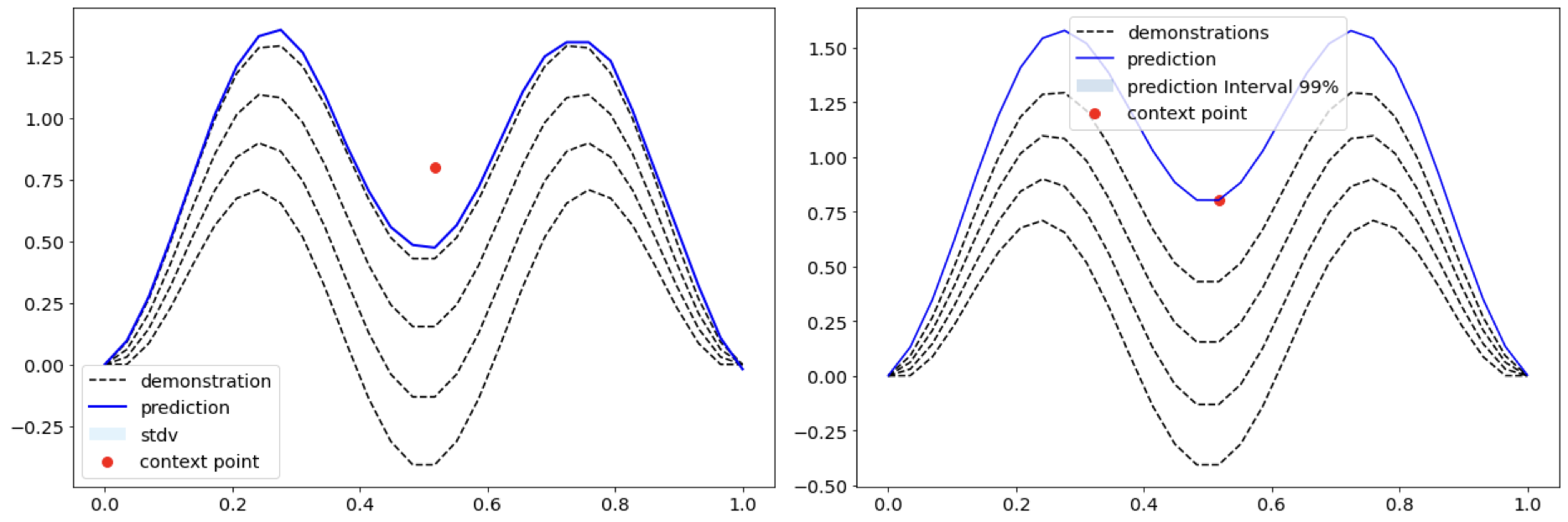}
    \includegraphics[width=\linewidth]{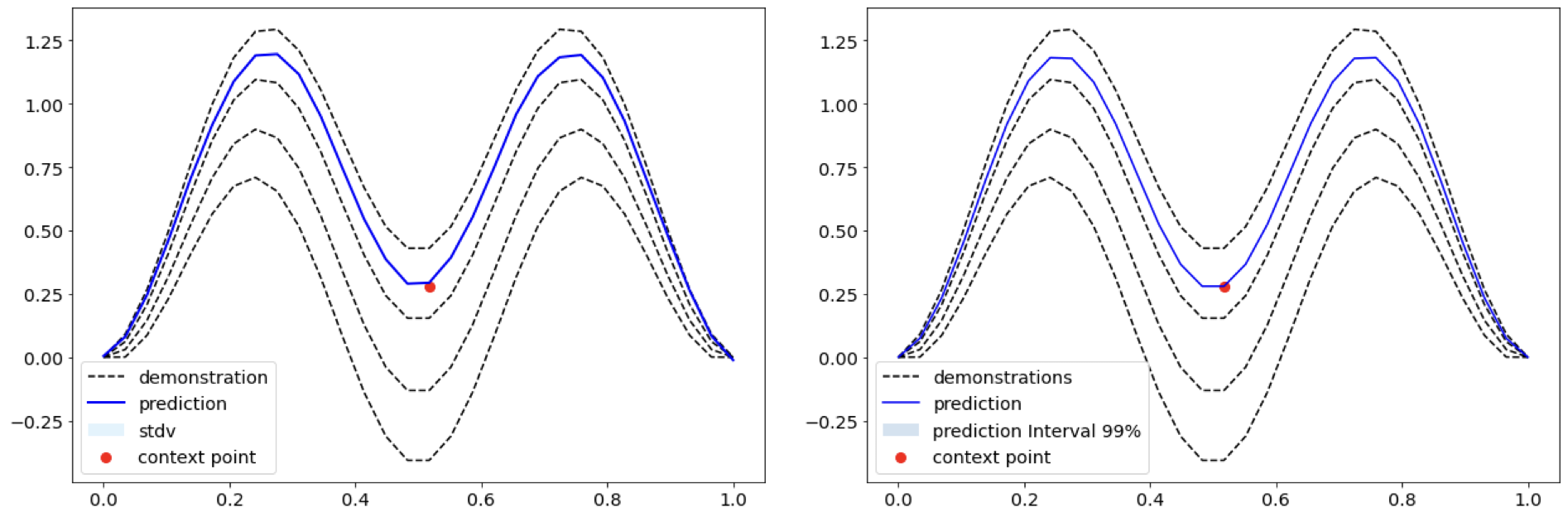}
    \caption[Comparison of estimated trajectories and confidence levels between CNMP and CESN+ models - single context]{Comparison of estimated trajectories and confidence levels between CNMP (left) and CESN+ (right) models.}
    \vspace{-5mm}
    \label{fig:cm1}
\end{figure}

As shown in Fig. \ref{fig:cm1}, the trajectory predicted by CNMP fails to satisfy the given condition, deviating significantly from the expected path. Furthermore, CNMP’s confidence metric is misleading, as it indicates high confidence despite producing a grossly inaccurate trajectory. In contrast, CESN+ aligns closely with the specified context, generating a trajectory that accurately satisfies the given condition. Moreover, CESN+ provides a consistent and dependable confidence metric, reflecting the true reliability of its predictions.

\subsection{Multiple context points}
We performed a similar task with multiple context points. This assessment aimed to determine not only the accuracy of the generated trajectories relative to the conditions but also the reliability of the models' confidence metrics under multiple conditions.

\begin{figure}
    \centering   
    \includegraphics[width=\linewidth]{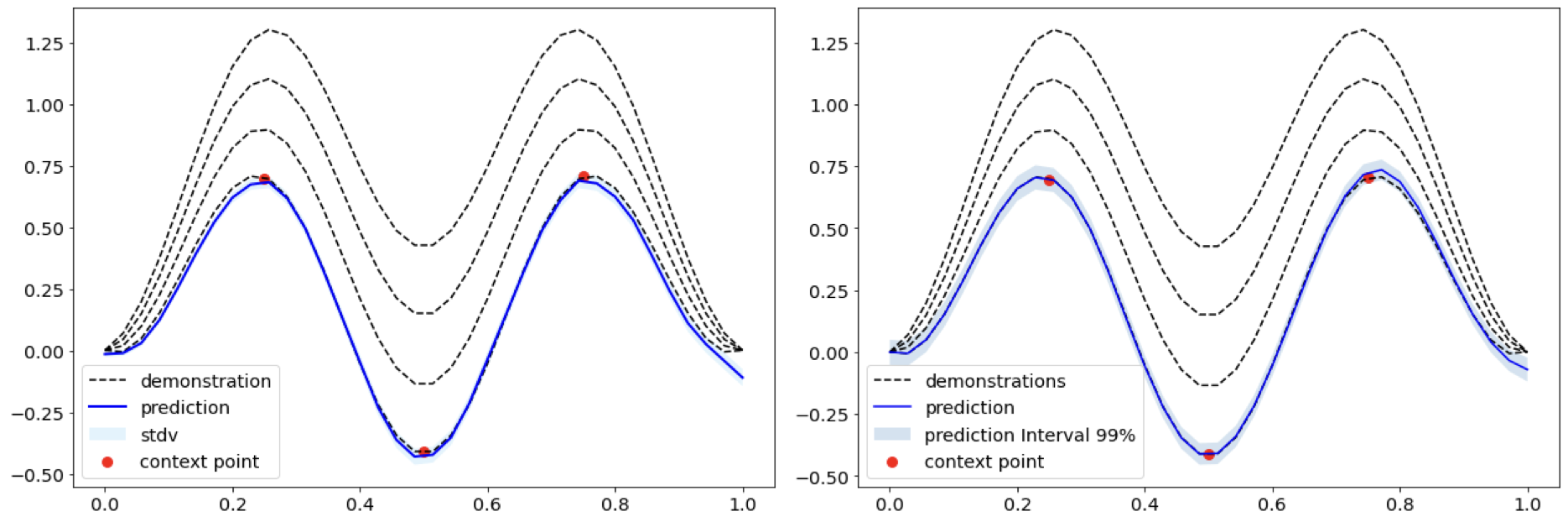}
    \includegraphics[width=\linewidth]{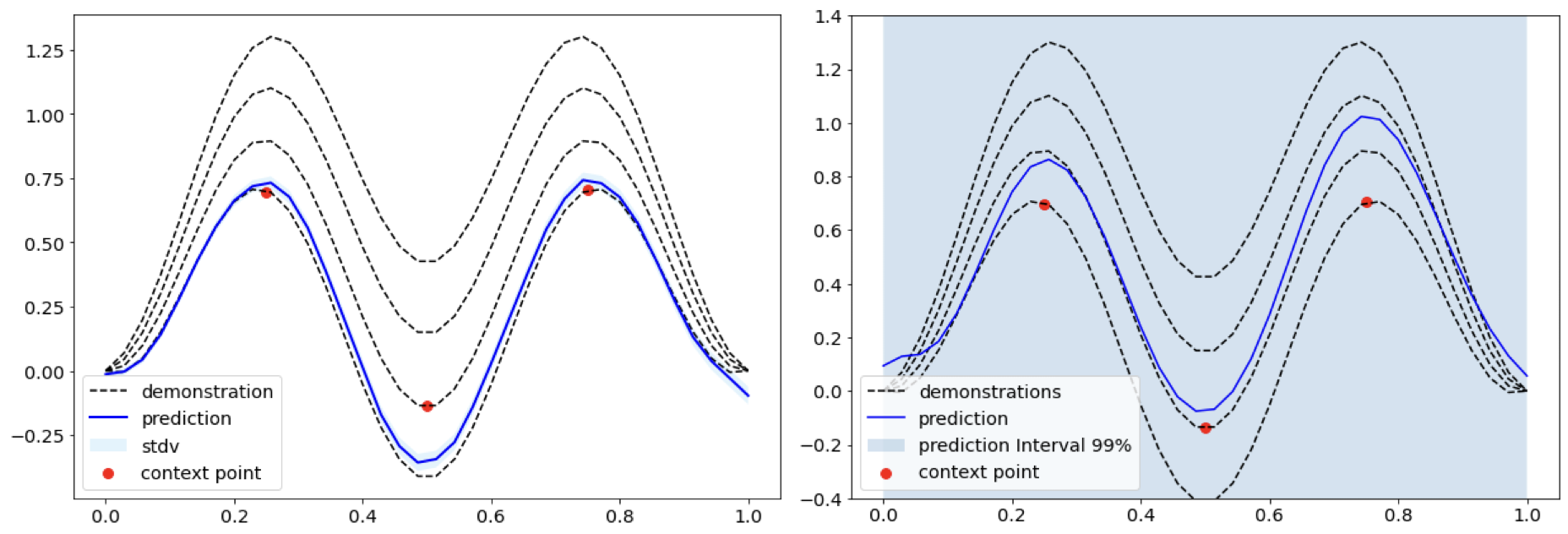}
    \includegraphics[width=\linewidth]{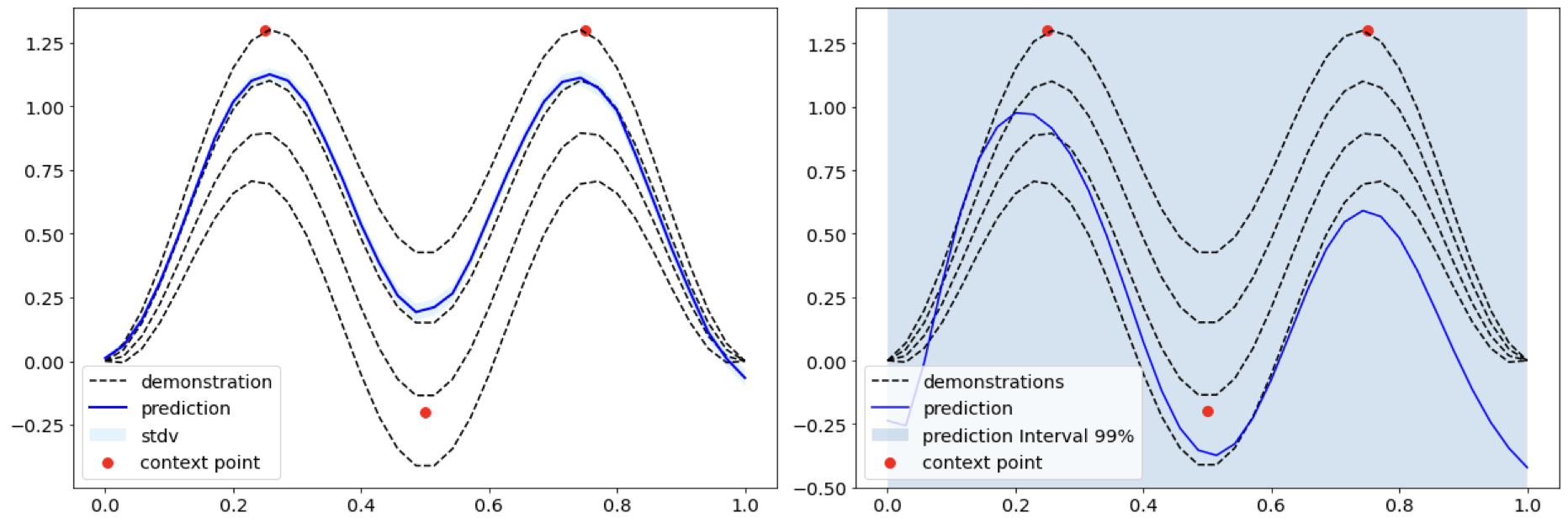}
    \caption[Comparison of estimated trajectories and confidence levels between CNMP and CESN+ models - multi context]{Comparison of estimated trajectories and confidence levels between CNMP (left) and CESN+ (right) models.}
    \label{fig:cm3}
    \vspace{-4mm}
\end{figure}

As illustrated in Fig. \ref{fig:cm3}, both models struggle to produce trajectories that align with the conditions specified by the context points. 
Despite this, the CNMP model exhibits excessive confidence, undermining the trustworthiness of its predictions in uncertain situations.
In contrast, the CESN+ model demonstrates correctly a low level of confidence in its estimates, reflecting its recognition of the inaccuracies in its predictions and providing a more reliable assessment of uncertainty.

\section{ROBOTIC EXPERIMENTS AND RESULTS}

To demonstrate the suitability of CESN+ for robotic adaptive shared control, we conducted a series of experiments showcasing how CESN+ can effectively generate context-dependent movement trajectories while providing reliable confidence metrics, essential for adaptive human-robot interaction. 

\subsection{Experiment Design}
\begin{figure}
    \centering   
    \includegraphics[width=\linewidth]{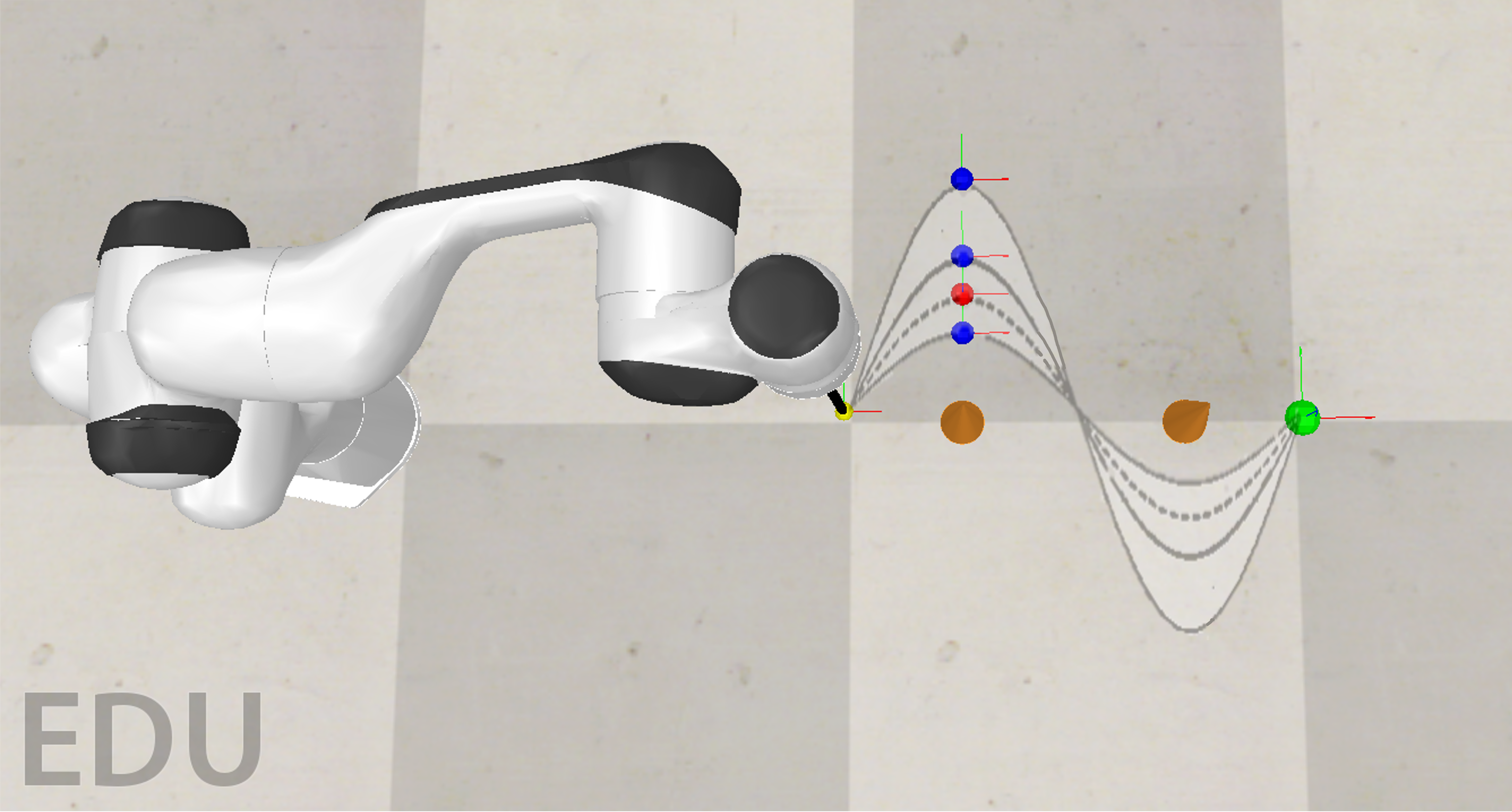}    
    \caption{Simulation scene in CoppeliaSim, illustrating the Franka Emika robotic arm,  equipped with an attachment }
    \label{fig:simu}
    \vspace{-2mm}
\end{figure}

In this experiment, we utilize the Franka Emika robotic arm, which features a 7-DOF (Degrees of Freedom) configuration. The arm is equipped with a lightweight attachment that enhances its ability to interact with the environment. The experimental scenario is simulated in CoppeliaSim, where two stationary obstacles are positioned along the arm's path to the designated goal. The primary objective of the task is to \erhan{move the robot endeffector  from its initial position to the goal while avoiding collisions with the obstacles.} The simulation scene is shown in Fig.\ref{fig:simu}.

\edit{This task is structured to compare two operational conditions; "Fixed Weight Sharing" and "Adaptive Weight Sharing", to assess the effectiveness of using the prediction confidence of CESN+ for adaptive shared control.}
The hypothesis underlying this comparison is that the adaptive sharing approach, guided by CESN+'s confidence metric, would demonstrate superior performance in scenarios where the robot needs to adjust its behavior based on the reliability of its predictions.

\subsubsection{Human Control}
In both conditions, the human operator sends control commands through a joystick, where the displacements along the x and y axes are scaled by a predetermined scalar value to achieve an intuitive control, while the position z remains constant throughout the task.\edit{The desired end-effector pose is used to compute the seven joint angles of the Franka Emika Panda robot using CoppeliaSim's inverse kinematics, operating at a control frequency of 10 Hz.}


\subsubsection{Robot Control}
\begin{figure}
    \centering   
    \includegraphics[width=0.9\linewidth]{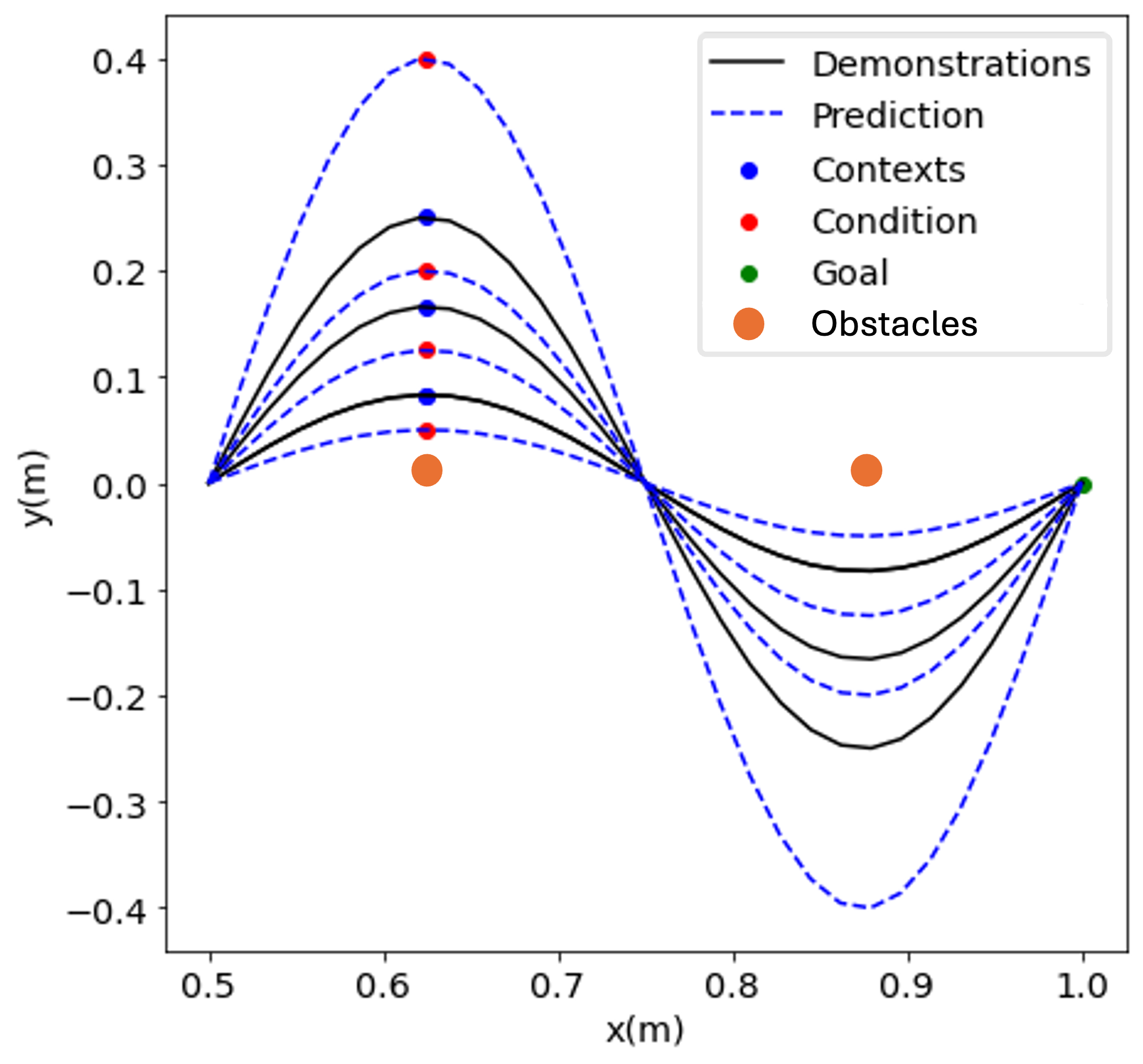} 
    \caption{Performance of CESN+ in generating robot trajectories. 
    \edit{"Contexts" refer to the respective contexts for each trajectory in the training dataset. In the prediction phase, "Condition" serve as context provided to the CESN+.}}
    \label{fig:perf}
\end{figure}

In this task, the robot autonomously generates control commands based on CESN+ predictions. The CESN+ is trained using three synthetic trajectories, each accompanied by its corresponding context. The trajectory generation is prompted by supplying the network with a desired context, which conditions the network to produce a trajectory that meets the specified conditions. For the CESN+, the reservoir size was set to 500. The performance of the CESN+ for both interpolation and extrapolation cases is illustrated in Fig. \ref{fig:perf}.

\subsubsection{Shared Control}
When the robot collaborates with a human to accomplish a task, it is important to account for the variability in the human’s chosen trajectories to reach the goal. To enhance the robot’s adaptability to human behavior, we can introduce multiple checkpoints throughout the movement where the robot’s prediction or decision can be fully updated. In this experiment, as a proof of concept and for simplicity in analysis, we implement this idea with a single checkpoint. When the end effector reaches this checkpoint, the robot captures its current state and uses this as a condition for CESN+ to predict an autonomous trajectory that aligns with that condition.

The shared controller operates under two distinct conditions:

\textbf{Fixed Weight Sharing:} Prior to reaching the defined checkpoint, the control is entirely assigned to the human. Once the checkpoint is reached, the robot generates a trajectory that corresponds to the checkpoint condition. From that moment onward, the control weight (\(\omega = 0.5\)) is equally shared between the human and the robot.

\textbf{Adaptive Weight Sharing:} Similar to the previous condition, the robot waits for the checkpoint to make its prediction. However, the control weight is adaptively adjusted based on the prediction interval value at each time step. The normalized value of the prediction interval is utilized to determine \(\omega\), which represents the human control weight.

For each of the above conditions, a total of 14 trials, including 10 interpolation and 4 extrapolation cases was conducted.

\begin{figure}
    \centering   
    \includegraphics[width=0.49\linewidth]{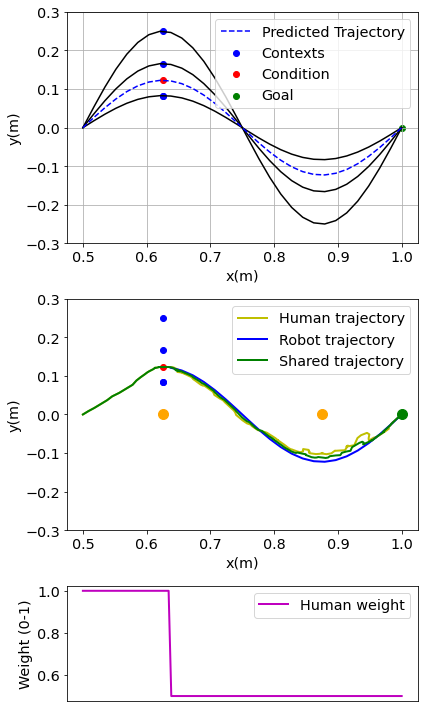}
    \includegraphics[width=0.49\linewidth]{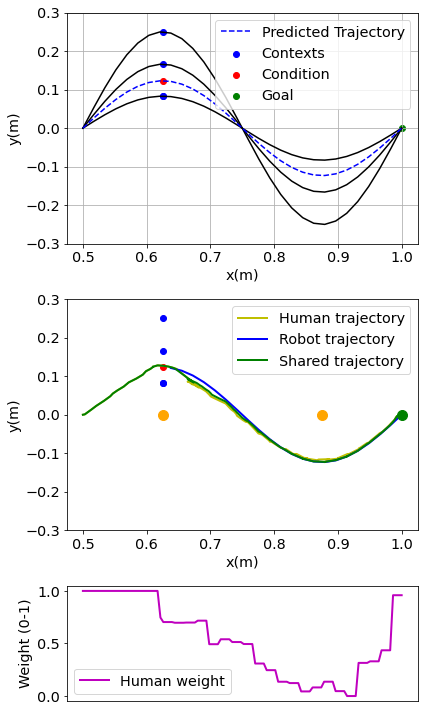}
    \caption{Task process of Fixed Weight Sharing and Adaptive Weight Sharing, demonstrating human desired vs robot's predicted vs the executed shared trajectory}
    \label{fig:fixvsadp}
\vspace{-4mm}
\end{figure}

\subsection{Results}
\begin{figure}
    \centering   
    \includegraphics[width=0.9\linewidth]{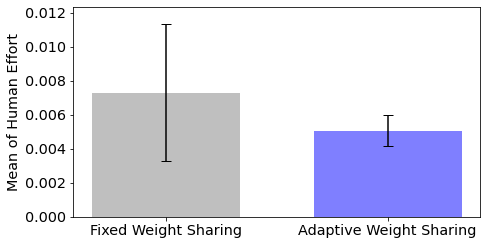} 
    \caption{Comparison of fixed and adaptive weight-sharing conditions. Error bars represent standard deviations.}
    \label{fig:effort}
\end{figure}

In this study, human effort is quantified as the absolute value of the human input vector at each time step, reflecting the level of control required from the human operator during the task. The analysis reveals significant differences in human effort between the fixed and adaptive conditions, as illustrated in the Fig. \ref{fig:effort} and detailed in the statistical analysis Table. \ref{tab:statistical_analysis}. This statistical analysis was conducted to assess the significance of differences in human effort between fixed and adaptive weight sharing control strategies.


To determine statistical significance, we used the two-sample t-test for normally distributed data and the Mann-Whitney U test as a non-parametric alternative. These tests confirmed that adaptive weight sharing significantly reduces human effort compared to the fixed approach. This would be an expected result if the uncertainty estimation and conditioning mechanisms of CESN+ indeed prove useful. Without such mechanisms, the performance with our model would not differ significantly from an uninformed 50-50 control sharing approach. This proof-of-concept indicates that CESN+ can be deployed for robotic shared control tasks where adaptive, confidence-based adjustments are essential for optimizing human-robot collaboration and reducing user workload.

\begin{table}
    \centering
    \caption{Statistical Analysis Results for Human Effort Comparison}
    \begin{tabular}{|l|c|c|}
        \hline
        \textbf{Test} & \textbf{Statistic} & \textbf{p-value} \\
        \hline
        Two-sample t-test & t = -5.8450 & $< 0.0001$ \\
        Mann-Whitney U test & U = 145784.0000 & $< 0.0001$ \\
        \hline
        \multicolumn{3}{|l|}{\textbf{Significance:}} \\
        \multicolumn{3}{|l|}{- Significant difference based on t-test ($p < 0.05$)} \\
        \multicolumn{3}{|l|}{- Significant difference based on Mann-Whitney U test ($p < 0.05$)} \\
        \multicolumn{3}{|l|}{- N for each category = 1176} \\
        \hline
    \end{tabular}
    \label{tab:statistical_analysis}
    \vspace{-4mm}
\end{table}

\section{CONCLUSIONS}

In this study, we introduced Context-based Echo State Networks with prediction confidence (CESN+) as a lightweight model for learning and online generation of robotic movement primitives suitable for human-robot shared control scenarios. CESN+ can be conditioned in real time to generate trajectories aligned with the desired trajectory of the human operator. We compared the performance of Conditional Neural Movement Primitives (CNMPs) with CESN+ for generating trajectories under single and multiple given conditions, where CESN+ demonstrated superior performance in both trajectory accuracy and reliability of prediction confidence. 

To demonstrate the usability of CESN+ for robotic shared control, a proof-of-concept study was conducted where CESN+ was applied to a robotic task, utilizing prediction confidence to adaptively adjust weight-sharing parameters for smoother collaboration and reduced human effort. The reliability of CESN+ was evident in its ability to accurately assess the confidence of its predictions, ensuring consistent and safe transitions between human and robotic control. To evaluate the effectiveness of adaptive weight sharing based on prediction confidence, we compared it with a fixed-weight scenario, and our results indicated a significant reduction in human control commands in the adaptive setting.

Future work could involve extending the approach to include multiple checkpoints throughout the trajectory, allowing for continuous updates to the robot’s predictions and decision-making. Another avenue for exploration is comparing CESN+ to other confidence-based models for adaptive control to provide a broader assessment of its performance. Additionally, implementing CESN+ in real-world robotic systems to evaluate its adaptability, reliability, and user experience under more complex and unpredictable conditions would be valuable for assessing its practical applicability.





\section{ACKNOWLEDGMENTS}
This work was supported by the project JPNP16007 commissioned by the New Energy and Industrial Technology Development Organization (NEDO) and by the Japan Society for the Promotion of Science KAKENHI Grant Number JP23K24926.
    
\bibliographystyle{ieeetran}
\bibliography{references}

\end{document}